\documentclass[letterpaper, 10 pt, conference]{IEEEconf}
\IEEEoverridecommandlockouts
\usepackage{cite}
\usepackage{amsmath,amssymb,amsfonts}
\usepackage{algorithmic}
\usepackage{algorithm}
\usepackage{tabularx}
\usepackage{hyperref}
\usepackage{graphicx}
\usepackage{textcomp}
\usepackage{xcolor}
\usepackage{verbatim}

\def\BibTeX{{\rm B\kern-.05em{\sc i\kern-.025em b}\kern-.08em
    T\kern-.1667em\lower.7ex\hbox{E}\kern-.125emX}}
\begin{document}

\title{Multi-Object Grasping -- Generating Efficient Robotic Picking and Transferring Policy
}

\author{Adheesh Shenoy, Tianze Chen, and Yu Sun
\thanks{The authors are from the Robot Perception and Action Lab (RPAL) of Computer Science and Engineering Department, University of South Florida, Tampa, FL 33620, USA. Email: \texttt{\{shenoy, tianzechen, yusun\}@usf.edu}. Adheesh Shenoy is an undergraduate student.}}
\maketitle

\begin{abstract}
Transferring multiple objects between bins is a common task for many applications. In robotics, a standard approach is to pick up one object and transfer it at a time. However, grasping and picking up multiple objects and transferring them together at once is more efficient. This paper presents a set of novel strategies for efficiently grasping multiple objects in a bin to transfer them to another.  The strategies enable a robotic hand to identify an optimal ready hand configuration (pre-grasp) and calculate a flexion synergy based on the desired quantity of objects to be grasped.  This paper also presents an approach that uses the Markov decision process (MDP) to model the pick-transfer routines when the required quantity is larger than the capability of a single grasp. Using the MDP model, the proposed approach can generate an optimal pick-transfer routine that minimizes the number of transfers, representing efficiency. The proposed approach has been evaluated in both a simulation environment and on a real robotic system.  The results show the approach reduces the number of transfers by 59\% and the number of lifts by 58\% compared to an optimal single object pick-transfer solution. 
\end{abstract}

\section{Introduction} 
Transferring multiple objects from one bin to another can be considered a menial task for humans. Using the sense of touch and experience, we can simply grasp multiple objects from a pile and move them to a bin. We face these types of tasks in multiple situations. When we cook, based on the recipe, we grasp and transfer multiple cloves of garlic into a pot. In logistics, workers are expected to transfer objects such as bulbs from a pile to fill 10-pack or 5-pack bins. In all these tasks, we grasp several objects at a time and transfer them together because it is more efficient than grasping and transferring an object one at a time. 

However, we have not observed many works on multiple-object grasping (MOG) or multiple-object bin-picking in robotics.
A significant amount of works have been focusing on single object bin-picking, pick-and-place, and grasping for manipulations. In traditional single object picking/grasping, an object's pose would be estimated using a vision system \cite{lowe1991fitting, dementhon1995model, zhu2014single, agrawal2010vision} to guide a robotic hand/gripper. Since humans have demonstrated outstanding grasping skills, several approaches extract grasping strategies and use them to reduce the complexity of grasp planning \cite{lin2012learning,huang2019dataset, lin2015robot,lin2014grasp}.  
More recently, deep-learning-based approaches use large labeled datasets and deep neural networks to directly find good grasp points from dense 3D point clouds \cite{mahler2017learning, 
lenz2015deep,ten2018using, kappler2015leveraging}. A comprehensive review can be found in \cite{li2019survey}.

Only a limited amount of work on grasping multiple objects has been carried out for static grasp stability analysis. \cite{harada1998enveloping} discusses the enveloping grasp of multiple objects under rolling contacts and  \cite{harada2002active} studied force closure of multiple objects. It builds the theoretical basis for later work on active force closure analysis for the manipulation of multiple objects in \cite{harada2002active}. \cite{yoshikawa2001optimization, yamada2005grasp, yamada2015static} try to achieve stably grasping of multiple objects through force-closure-based strategies. In their studies, the target objects are already in the air and traditional grasp quality measures were used to analyze the grasps \cite{lin2015grasp,LinISRR2013,lin2015task,Sun2016}. Our recent work \cite{chen2021iros} has studied the tactile sensing aspect of MOG and developed a deep learning approach to estimate the object quantity in the grasp. 

There are several technical challenges in MOG.  Firstly, estimating the object quantity and their pose in a bin is very challenging. Occlusion among objects of similar color and texture makes a computer-vision-based approach prone to error. For example, in the 2019-2020 IROS Robotic Grasping and Manipulation Competition (RGMC) \cite{sun2021robotic}, all teams failed at picking ice cubes from an ice bucket. Secondly, The displacement of objects within the bin in contact with the hand voids previously estimated poses. When the hand makes contact with the object pile in the bin, it displaces the objects. If we only rely on a computer vision system, the eye of the hand-eye system can no longer view the majority of the hand. Therefore, it can no longer update the estimated pose of the objects to be grasped. 

A robot will have to use tactile sensors and torque sensors when grasping objects from a bin. Tactile sensing has widely been recognized as a critical perception component in object grasping and manipulation. It has been used with vision sensors to estimate the location of an object relative in a world coordinate system \cite{chhatpar2005particle, chebotar2014learning}, embedded force sensors on a robotic hand  \cite{corcoran2010measurement}, and a 6-axis  force/torques sensor on a robot’s wrist \cite{petrovskaya2006bayesian, javdani2013efficient, bimbo2015global, petrovskaya2011global, saund2017touch}. In those works, tactile/force sensors are used to reduce the uncertainty in the perception of the vision system or are only used for single object grasping.  
For picking the desired quantity, the robot would need to predict how many objects will remain in grasp after lifting the hand out of the bin. The robot needs to make the prediction before lifting the hand so that it can adjust or simply try again without lifting the hand if the predicted quantity is different from the desired one.

When the desired quantity is small, MOG once could be sufficient. However, if the desired quantity is large, one grasp and transfer would not produce it.  We would need to develop an approach that can produce a large quantity MOG policy to ensure both the precision of the total outcome of several MOGs and transfers, and the efficiency of the combined MOGs and transfers.  

The contribution of this paper includes two novel procedures for transferring a targeted quantity of objects from a pile into a bin using MOG. It also introduces several new techniques that are designed for MOG, including a pre-grasp selection, end-grasp selection, maximum capability grasp selection, and in-grasp object quantity estimation. Compared to single object transferring, our approach reduces the total number of transfers between bins by 59\% and the number of lifts from the bin by 58\% assuming that the single object grasping algorithm has a 100\% success rate.

\section{Problem Description and Approaches}
\label{sec-problem}

\begin{figure}[htbp]
    \centering
    \includegraphics[width=0.4\textwidth, height=3.5cm]{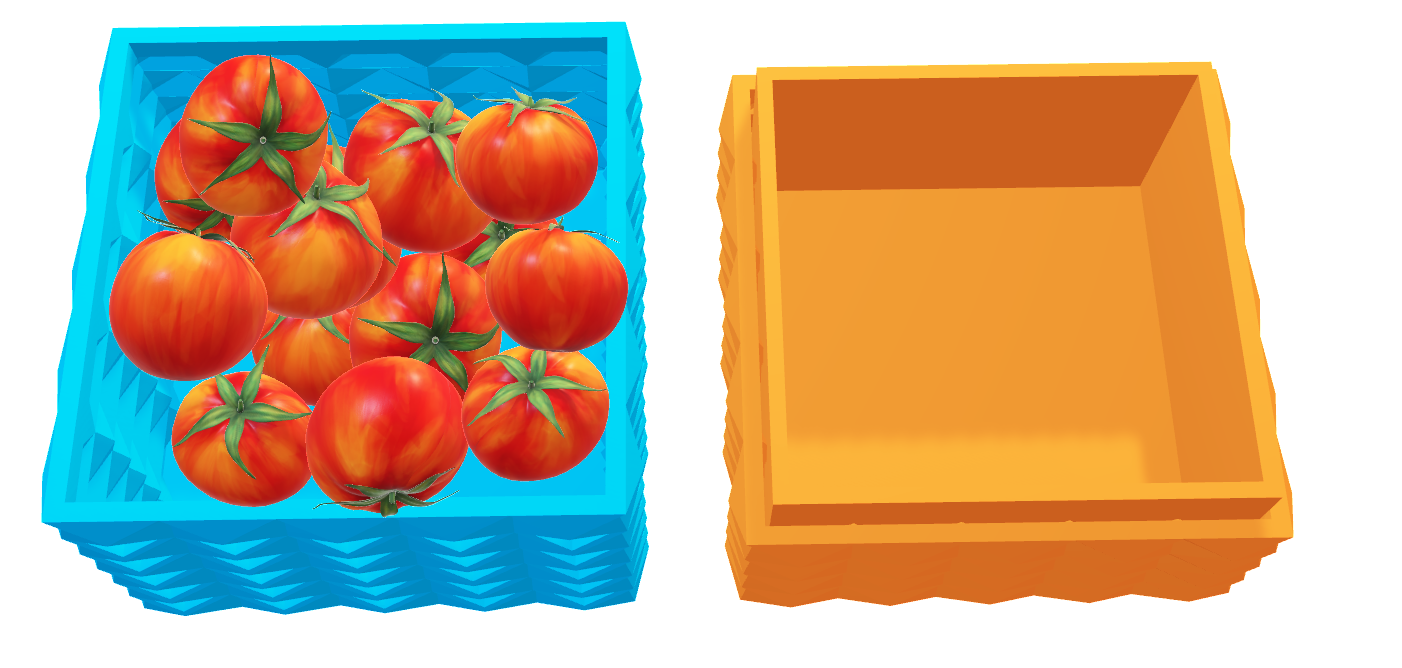}
    \caption{A multi-object grasping and transferring example. A robot is asked to transfer 10 tomatoes from the blue bin to the yellow bin. }
    \label{fig-bin}
\end{figure}

In this paper, we focus on transferring a large quantity of uniformly shaped and sized objects from one bin to another. An example setup is illustrated in Figure \ref{fig-bin}.  The objects in the original bin are randomly piled up.  The pick-and-transfer process for a large quantity can be solved in several different ways. This paper compares three different approaches.  The first approach is the basic single-object grasping approach to pick and transfer one object at a time.  If the quantity is $N$, it would need at least $N$ times of picking and transferring. 

The second approach is a naive multi-object grasping approach that first grasps as many objects as possible to quickly reach or get close to the desired target quantity and then grasps the remaining number. If a robot can grasp and hold $q$ objects at the most, the robot will first perform $p = rounddown(\frac{m}{q})$ times of picking and placing $q$ objects and then grasp the remaining $r=N-p*q$ number of objects.  For example, if the demanded quantity is $10$ and a robotic hand can grasp $4$ objects at the most, the robot will grasp and transfer $4$ objects twice and then pick and transfer the remaining $2$ objects.  
However, in reality, even if a robotic hand can grasp $4$ objects at the most, it can rarely do it successfully and consistently. It may need many re-grasps (open and close the hand in the pile) to achieve it.  On the other hand, because of the perception error, a robot may think it holds $4$ objects and lifts the hand,  but it may have grasped $1$ or $2$ or even $0$ objects in the hand.  Therefore, the naive multi-object grasping approach may not be the most efficient approach for those two reasons.

This paper introduces another approach that models the picking and transferring process as a Markov decision process (MDP) because of the stochastic feature in grasping multiple objects. As illustrated in Figure \ref{fig-bin}, the states are the object quantity in the receiving bin and the actions are grasping actions for a different number of objects. Since the grasping action may pick up different quantities with different probabilities, it is an MDP. The optimization goal is to reach the desired quantity while minimizing the number of grasps and transfers between bins.  The model may generate a policy that requires the robot to perform a grasp action for any number of objects at a particular step.  We call this approach the \emph{Markov-decision-process-based multi-object grasping and transferring or MDP-MOGT}.  

\begin{figure}[htbp]
    \centering
    \includegraphics[width=0.5\textwidth, height=4.5cm]{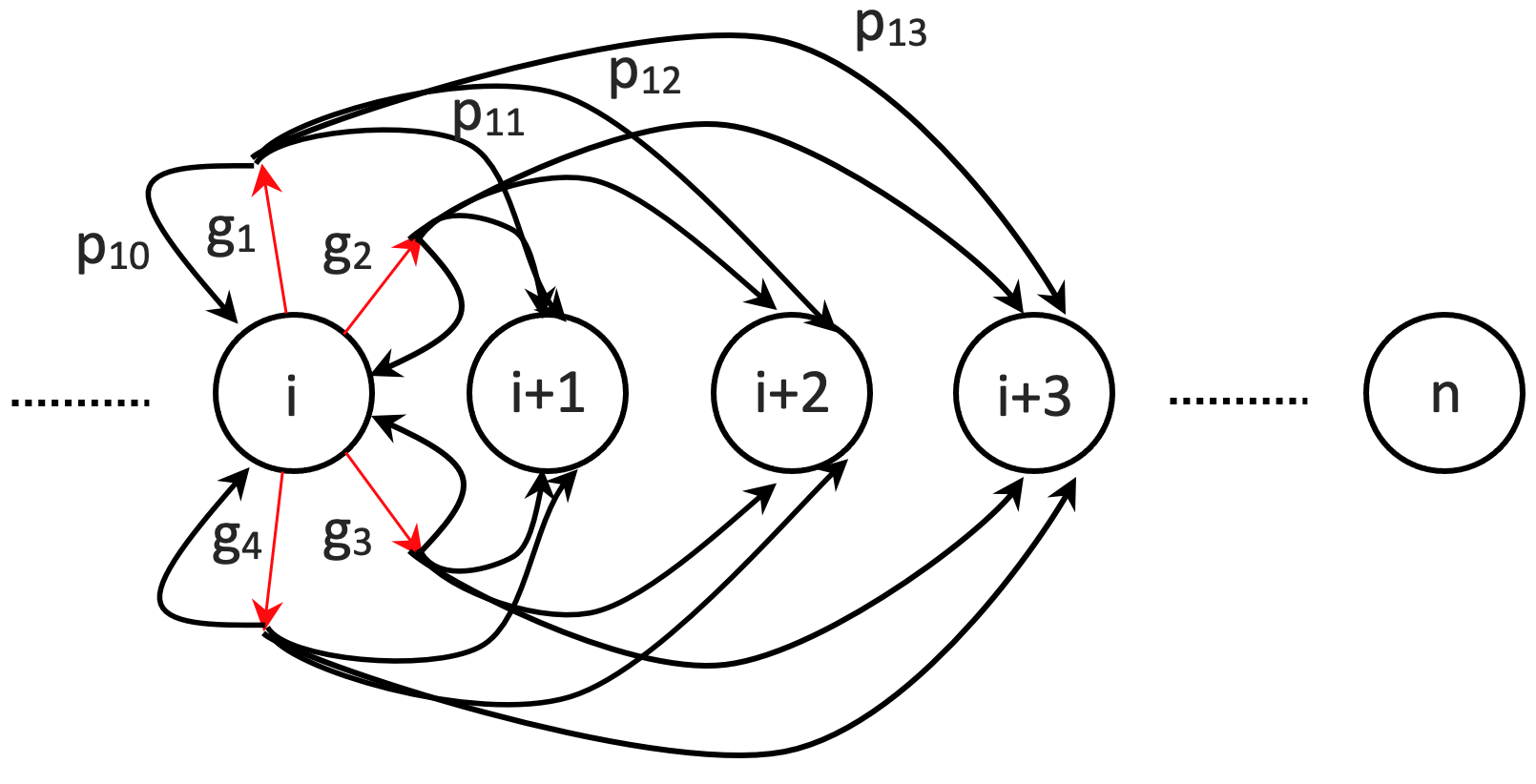}
    \caption{High quantity grasping and transferring process is modeled with a MDP. $0$, $1$, ... $n$ in the circles are states.  They represents the number of objects in the receiving bin.  $g_i$'s are MOG actions. Each MOG action $i$ could end up having $0$, $1$, $2$, ..., or $q$ objects in the grasp and transit to state $i$, $i+1$, ..., or $i+q$ with $p_{i0}$, $p_{i1}$, ..., $p_{iq}$ probabilities respectively.  }
    \label{fig-mdp}
\end{figure}

\section{Markov-decision-process-based multi-object grasping and transferring}
\label{sec-mdp}


\subsection{State space and rewards}
The state is the object quantity in the receiving bin. If the receiving bin has the desired quantity, we reach the goal state.  The start state is zero that represents an empty bin.  At a given step, a robot takes an action of MOG and transfers objects to the receiving bin if there are objects grasped. At the end of each step after the robot has taken an action, the state of the robot will change.  As illustrated in Figure \ref{fig-mdp}, the states are from $0$ to $n$, where $n$ is the desired quantity.

The reward is associated with the new state. We define the reward as follows:

\begin{equation} \label{reward}
r(s^{'}|(s,a)) =  
\begin{cases}
    -(n - s^{'}),         & \text{if } s^{'} < n\\
    100,000             & \text{if } s^{'} = n\\
    -1000,                & \text{if } s^{'} > n\\
\end{cases}
\end{equation}

where $r(s^{'}|(s,a))$ is the reward of taking action $a$ at state $s$ resulting in the new state $s^{'}$. 


\subsection{MOGT action space}
\label{sec-action-space}
A MOGT action contains three sub-actions: MOG, lifting, and transferring.  Any sequence of finger flexion and extension could be a grasping action.  However, some sequences can rarely pick up any objects, while a small few can grasp multiple objects with a high success rate.  Therefore, first, we must explore the grasping action space and identify several grasps that are suitable for multiple object grasping.  

\begin{figure}[htbp]
    \centering
    \includegraphics[width=0.8\linewidth]{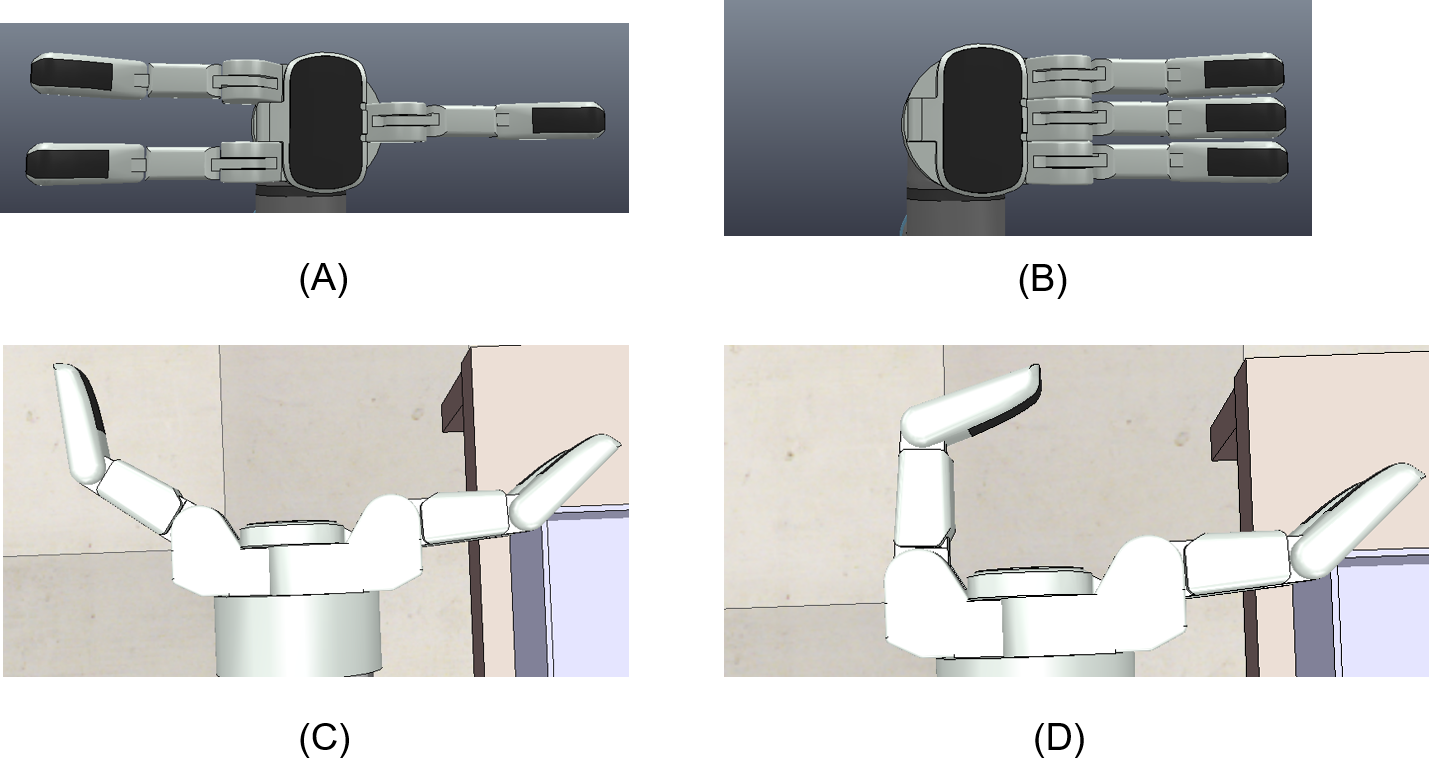}
    \caption{(A) Spread is at $0^{\circ}$. (B) Spread is at $360^{\circ}$. (C) Fingers on the left side are at $30^{\circ}$. (D) Fingers on the left are at $90^{\circ}$.}
    \label{fig-pre_grasp_sampling_example}
\end{figure}

To fully explore the grasp action space, we have developed a stochastic flexing routine (SFR) in \cite{chen2021iros} and used it to perform a bias random walk grasp from several pre-grasps. The pre-grasps set has a dense sample of feasible hand configurations.  In our experiment, we use uniform sampling to obtain 9,000 pre-grasps for a Barrett hand. Figure \ref{fig-pre_grasp_sampling_example} shows the upper and lower bounds of the spread angle and finger angle that form the pre-grasp set. For the set, we choose $20^{\circ}$/step for the spread angle and $3^{\circ}$/step for the finger base joint. These step sizes were chosen to reduces the sample size, yet be extensive enough for our search. 

\subsubsection{Selecting pre-grasps based on potentials}
\label{sec-techniques-potentials}
Not all pre-grasps lead to good MOGs.  To pick the best ones, the robot would perform SFR $10$ times from every pre-grasp in the pre-grasp set. We use their grasping results to calculate the potential pre-grasp value (PPG). 

\begin{equation} \label{PPG}
PPG(\mathbf{\theta}, \bf{O}) = \{p_0, p_1, p_2, ..., p_m\},
\end{equation}
where $\boldsymbol{\theta}$ is the robotic hand's joint angle vector, $\bf{O}$ is the object's geometry model, $p_i$'s are the probabilities of the pre-grasp leading to a successful grasp of $i$ objects.

We compare the average $PPG$ of each spread to define the best spread for grasping the target quantity of objects. We filter the pre-grasps within that spread based on their success rate for grasping the target quantity of objects and apply K-mean clustering on the remaining data to obtain clusters of hand configurations. 

We perform the SFR on the centroids of the clusters 100 times and compute the $PPG$ for the centroids. Finally, we select the best pre-grasp from the centroids based on its $PPG$ to give us the clustered-probability-based pre-grasp (CPPG) for each targeted grasp

\subsubsection{Selecting pre-grasps based on expectations}
\label{sec-expectations}
We can also find the pre-grasp that can successfully transfer a large quantity of objects at once based on the best expectation pre-grasp (BEPG). We define the BEPG as the pre-grasp that, on average, yields the highest quantity of objects. The strategy we propose to acquire the BEPG also relies on the pre-grasp set mentioned in section \ref{sec-techniques-potentials}.

We use the $PPG$ to calculate the average grasp potential (AGP) or grasp number expectation for each spread within the pre-grasp set. 
\begin{equation}\label{AGP}
AGP(\mathbf{\theta}, \bf{O}) = p_1+p_2*2+...+p_i*i + p_m*m,
\end{equation} 
where the AGP is the weighted sum of the quantity of grasped objects and the weights are their probabilities.
Therefore, we select the spread with the best average AGP. We filter the pre-grasps within that spread based on their AGP and apply K-mean clustering to obtain clusters of hand configuration that have a high potential for grasping large quantities of objects.

Similar to the strategy in section \ref{sec-techniques-potentials}, we perform the SFR on the centroids of the clusters 100 times and select the pre-grasp with the highest AGP value. This pre-grasp would be the BEPG.

\subsubsection{Selecting pre-grasps based on volume}
\label{sec-techniques-MCPG}
When transferring a large quantity of objects between bins, humans usually grasp a \textit{handful} of objects in each attempt. We try to expand our hands as large as we can and attempt to grasp the maximum quantity in each transfer. We call this pre-grasp, the maximum capability pre-grasp (MCPG). We believe that during the preliminary transfers between bins, the robot hand should grasp as many objects as possible. One approach for finding this pre-grasp is to compute the volume of the in-grasp space of the hand. Therefore, we can use the grasp with the largest volume to get the MCPG for a hand. 

We use the strategy mentioned in \cite{chen2021iros} to compute the volume of every pre-grasp in the pre-grasp set. We can use the pre-grasp with the largest volume as the MCPG for the hand. 

\subsubsection{Selecting finger flexion synergy}
\label{sec-techniques-types}
Using SFR, for a single pre-grasp we discovered several distinct end-grasp types associated with the quantity of objects in the hand. We define the end-grasp as the hand configuration when the grasping routine has been completed. Based on our observation, some end-grasps have a higher chance of grasping a particular quantity of objects compared to others. Therefore, to compare the end-grasps we define a success rate for the grasp type with

\begin{equation} \label{SRG}
    SRG(k, \bf{O}) = \{s_0, s_1, s_2, ..., s_m\},
\end{equation}
where $k$ is the index of a grasp type, $s_i$ is the success rate that the grasp type $k$ has in grasping $i$ objects. Therefore, a end-grasp type can be selected based on its $SRG$ values. 

Using the data collected, we apply K-mean clustering on the end-grasps to obtain clusters of hand configurations that are likely to fit the desired quantity of objects. We use the centroids with the highest neighbors as the end-grasp. This end-grasp is used to compute the finger flexion synergy. 

\subsubsection{Lifting based on prediction} 

\label{sec-techniques-estimate}
Before the robot lifts the hand, it must sense the objects in the grasp and predict how many objects will remain in the grasp after lifting. It is primary for successful MOG and consequently, successful transfer of objects between bins.  It is part of the MOG action because the robot will continuously re-grasp until the robot senses that the desired number has been reached and it can lift the hand from the original bin. 

Since modeling the physics of the objects within the bin is computationally expensive, we decided to develop a data-driven deep learning approach to estimate the quantity of objects within the grasp when the hand is inside the bin.

\begin{figure*}[htbp]
    \centering
    \includegraphics[width=0.8\textwidth, height=6cm]{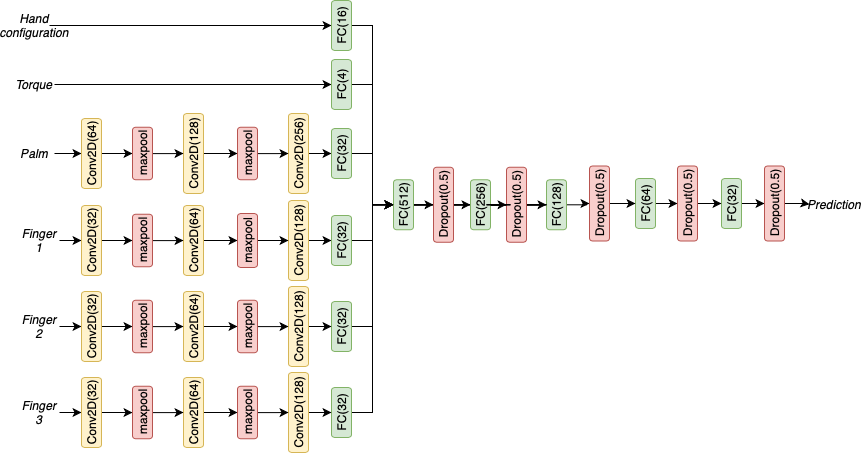}
    \caption{$hand\ configuration$ is a vector that contains the pre-grasp and the current hand configuration. $torque$ is a vector that contains the strain gauge readings from each finger. $palm$ represents the tactile sensors on the palm of the hand and $finger1$ to $finger3$ represents the tactile sensor array on each finger. $prediction$ is the classification result.}
    \label{fig-classifier}
\end{figure*}

We train five classifiers that estimate zero or non-zero objects, two or non-two objects, three or non-three objects, and greater than or equal to 2 objects or lesser than 2 objects in the hand. The structure of the classifier is shown in Figure \ref{fig-classifier}. The input data dimension of the model is $113$. The input data contains the pre-grasp and the current hand configuration ($14$), the tactile sensor information ($96$), and the readings from the strain gauges present in the couple joints of the hand ($3$). The activation function used for the output layer is sigmoid with a single class as the output. The whole model can be represented as  
\begin{equation} \label{classifier}
n = f(\mathbf{h}, \mathbf{t}, \mathbf{s}),
\end{equation}
where $\mathbf{h}$ is a vector representing the hand configuration, $\mathbf{t}$ is the vector representing the tactile sensor array, and $\mathbf{s}$ is the vector containing the three strain gauge readings. The output $n$ is the prediction of whether the grasp would contain the desired quantity of objects.
The shape for the tactile sensor array is rearranged to account for the spatial information. We represent the tactile sensors in the palm as a $4$ by $7$ matrix and the tactile sensors in each finger as a $3$ by $8$ matrix. The result of the prediction models can be found in section \ref{sec-results-model} 

To reduce the number of false positives when performing the grasping routine, we use the non-zero model along with either of the models to estimate when to lift the hand. If the non-zero model estimates non-zero for three consecutive time steps and the other model estimates True for one time-step we lift the hand. We define this as the voting algorithm. We use the $\geq 2$ model when grasping the most quantity of objects and the $1$, $2$, and $3$ object models when we grasp their respective quantity of objects.

\subsubsection{Transfer sub-action}
The last part of the MOGT action is the transfer sub-action. 
If the number of objects in the grasp after lifting is desirable, we transfer them to the destination bin. If it is not, we drop the objects in the origin bin and repeat the grasping routine. 
For the sake of this paper, we assume that we can detect the quantity of objects within the hand once it has been lifted from the pile. 

\subsection{State transition probability}
After a MOGT action, a number of objects could be added to the receiving bin. So the system will transit to a new state with a probability. The actions we focus on are to grasp a maximum quantity of objects, $1$ object,  $2$ objects, and $3$ objects.
We obtain the state transition probability distribution for each action through data collection. 

\subsection{Obtaining and Applying MDP-MOGT Policy} 
\label{sec-stochastic-transfer}

We utilize the value iteration method based on the Bellman equation in equation \ref{eq-bellman} to compute the optimal policy when we are at each state of the MDP-MOGT model.  After we have got the policy we will follow the algorithm in Algorithm \ref{Alg - transferring_algorithm}. 

\begin{equation} \label{eq-bellman}
V_{t + 1}^{*}(s) = \max \sum_{s^{'}}p(s, a, s^{'})[r(s, a, s^{'}) + \lambda V_{t}^{*}(s)]
\end{equation}

\begin{algorithm}[h]
\caption{MDP-MOGT algorithm}
\label{Alg - transferring_algorithm}
\begin{algorithmic}[1]
\STATE Initialize $target$, $current$ as 0, $objectsGrasped$ as 0 
\WHILE{$current \ \neq \ target$}
    \STATE Initialize $isLift$ as 0
    \STATE choose pre-grasp Type following policy $\pi$
    \STATE choose corresponding end-grasp Type
    \STATE insert hand and start grasping
    \STATE $isLift$ = output from voting algorithm
    \IF{$isLift \ = \ 1$ }
        \STATE lift
    \ELSE
        \STATE reset hand to pre-grasp and regrasp
    \ENDIF
    \STATE lift and check number of objectGrasped
        \IF{$current + objectsGrasped <= target$}
            \STATE $current += objectsGrasped$
            \STATE transfer objects to destination bin
        \ENDIF
\ENDWHILE
\end{algorithmic}
\end{algorithm}

\section{Experiments and Results}
\label{sec-results}
We have evaluated our proposed approach in simulation as well as the real world using spheres. The robotic hand used in the evaluation is the Barrett Hand. Based on our data collection, the hand is capable of grasping a maximum of 5 objects, therefore, we use a target of $10$ objects, to showcase the advantage of high quantity transferring using MOG.

The two metrics we use for evaluating the entire approach are the number of transfers made between the two bins and the total number of lifts performed from the pile. These two metrics highlight the efficiency of the transfer approach.

\subsection{System setup}
\begin{figure}[htbp]
    \centering
    \includegraphics[width = 3cm, height= 3cm]{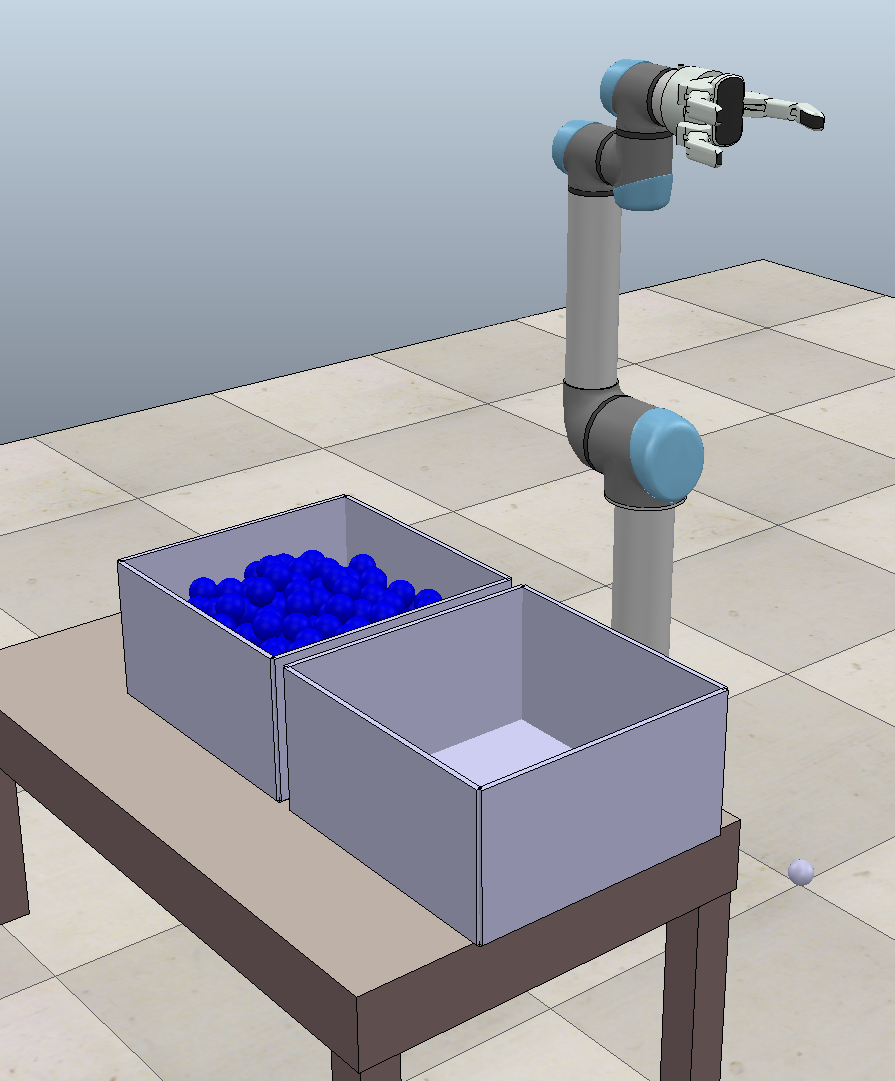}
    \includegraphics[width = 3cm, height= 3cm]{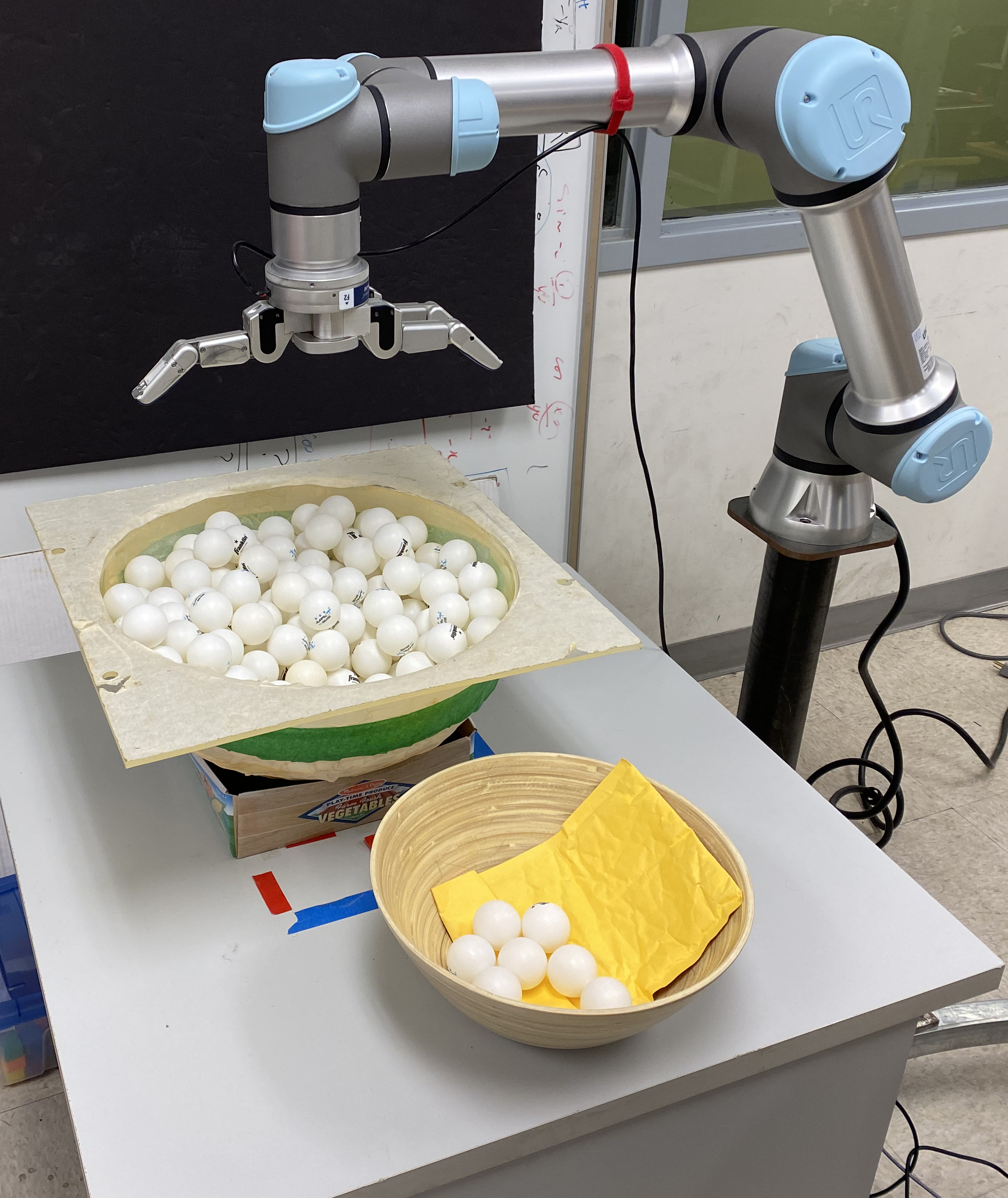}
    \caption{{Setup of our system in CoppeliaSim (left) and real system (right)}}
    \label{fig-set_up}
\end{figure}

The system for data collection was set up within simulation and the real world. We used CoppeliaSim as the simulation software. The setup within the simulation is shown in Figure \ref{fig-set_up}. We attach the Barrett hand to the UR5e robot arm and place the objects in a bin in front of the robot. The Barrett hand model provided in CoppeliaSim only contains one tactile sensor on each tactile sensing region. Therefore, we attached tactile sensors to the region to replicate the Barrett hand within the real system. The setup for the real system is also shown in Figure \ref{fig-set_up}.

We used spheres in both the simulation and the real system for our experiments. The spheres in both the systems have a radius of $2cm$ and a weight of $2.7g$.

\subsection{Data collection}
\label{sec-data}
\subsubsection{Data collection for CPPG and BEPG}
To gather the data for computing the pre-grasps, we used the simulation and the real system.

In the simulation, we repeat each of the $9000$ hand configurations from the pre-grasp set $10$ times using SFR.
For CPPG and BEPG, we use the data as described in sections \ref{sec-techniques-potentials} and \ref{sec-expectations} respectively to get the best pre-grasps. For computing CPPG, we focus on selecting the pre-grasps for grasping $1$, $2$, and $3$ objects. For selecting the number of clusters within each section, we visualize the inertia and distortion of the data and select the number of clusters where the inertia and distortion are decreasing linearly. Inertia can be defined as the sum of the squared distances of the data sample to their closest centroids. Distortion can be defined as the average of the squared distances from the centroid of each cluster.

For collecting data in the real system, we used the top 3 pre-grasps for the spheres from the simulation data for both CPPG and BEPG. We then performed SFR using those pre-grasps $50$ times using the real system to select the best pre-grasp for each grasp type for the real system.

\subsubsection{Data collection for the finger flexion synergy}
For getting the end-grasp for the pre-grasps within the simulation, we used the data collected within the simulation for the pre-grasps and use the approach in section \ref{sec-data}. We use inertia and distortion to select the clusters here as well.
For getting the end-grasp for the grasps in the real system, we used the data collected within the real system for each of the real system pre-grasps and perform the same approach as we do for getting the simulation end-grasps. 

\subsubsection{Data collection for MDP-MOGT}
\label{sec-state-transition-collection}
For the MDP-MOGT approach, we need to possess the state transition probabilities for each of the pre-grasps in the action space. Therefore, to get the state transition probabilities for the pre-grasps within the simulation, we used each pre-grasp and its corresponding end-grasp along with the estimation model to grasp the spheres $50$ times. This data provides us with the state transition probabilities for each pre-grasp when performing the grasping routine with the model and the finger flexion synergy. We also computed the state transition probabilities for each pre-grasp with SFR. We performed the same procedure in the real system to collect data and compute the state transition probabilities for the real system pre-grasps.

\subsection{Results of grasping maximum quantity of objects}
\label{sec-results-handful}
For grasping the maximum quantity of objects, we described two approaches in section \ref{sec-expectations} and \ref{sec-techniques-MCPG}, BEPG, and MCPG. We performed each grasp type $10$ times using SFR on the spheres in the real system. The results can be found in table \ref{tab-agp-volume}. The percentages within the table represent the percentage of times the pre-grasp was able to grasp the corresponding number of spheres. Based on the results, BEPG significantly outperforms MCPG in yielding more quantity of objects on average. 

\begin{table}[h]
\caption{best-expectation pre-grasp vs. maximum capability pre-grasp}
\label{tab-agp-volume}
\begin{center}
\begin{tabular}{|c|c|c|}
\hline
\textbf{\begin{tabular}[c]{@{}c@{}}Number\\of spheres\end{tabular}} &
\textbf{\begin{tabular}[c]{@{}c@{}}BEPG \end{tabular}} & 
\textbf{\begin{tabular}[c]{@{}c@{}} MCPG\end{tabular}}\\
\hline
0 & 6\% & 80\% \\
\hline
1 & 16\% & 10\% \\
\hline
2 & 16\% & 10\% \\
\hline
3 & 42\% & 0\% \\
\hline
4 & 20\% & 0\% \\
\hline
\end{tabular}
\end{center}
\end{table}

\subsection{Results of estimation models}
\label{sec-results-model}
We train a total of $5$ estimation models. The models are for estimating if the grasp contains at least 1 object (non-zero), $1$ object, $2$ objects, $3$ objects or at least $2$ objects. The precision and the RMSE for each model can be found in table \ref{tab-model-precision-RMSE}. Precision represents the percentage of the correct predictions among all predictions. RMSE represents the standard deviation of the prediction error.

For training each model, we use early stopping on the validation loss to prevent over-fitting. We also use the Adam optimizer with a learning rate of $0.001$ and use binary cross-entropy as our loss function. We apply SMOTE \cite{chawla2002smote} on the training data to fix the imbalanced classes. Lastly, we perform transfer learning using the real-system data to acquire the models for the real system. 

Based on the results, the precision for all the models reduces when performing transfer learning, except the model trained to estimate $\geq 2$ objects. We believe the reduction in the precision is because of the noise within the real system data. The reduction can also be attributed to limited real system data. 

\begin{table}[h]
\caption{Precision and RMSE of estimation models for sphere}
\label{tab-model-precision-RMSE}
\begin{center}
\begin{tabular}{|c|c|c|c|c|}
\hline
\textbf{\begin{tabular}[c]{@{}c@{}}Estimation \\ model\end{tabular}} & 
\textbf{\begin{tabular}[c]{@{}c@{}} Precision\\(Simulation) \end{tabular}} & 
\textbf{\begin{tabular}[c]{@{}c@{}} Precision\\(real) \end{tabular}} &
\textbf{\begin{tabular}[c]{@{}c@{}} RMSE\\(Simulation) \end{tabular}} & 
\textbf{\begin{tabular}[c]{@{}c@{}} RMSE\\(Real) \end{tabular}}\\
\hline
non-zero & 95.1\% & 96.09\% & NA & NA\\
\hline
1 & 85.97\% & 51.72\% & 0.6 & 0.69\\
\hline
2 & 67.18\% & 51.46\% & 0.97 & 0.87\\
\hline
3 & 40.95\% & 38.24\% & 1.84 & 0.92\\
\hline
$\geq 2$ & 78.97\% & 83.04\% & NA & NA\\
\hline
\end{tabular}
\end{center}
\end{table}

\subsection{Results of transfer approach}
For transferring objects between the bins, we have evaluated two approaches, the naive transfer approach, and the proposed MDP-MOGT approach. For MDP-MOGT, we collected data to compute the state transition probability for each pre-grasp and defined the problem as an MDP to acquire an optimum policy. 

For the experiments on both approaches, we used the BEPG as the pre-grasp to grasp the maximum quantity of objects and the CPPG for grasping the target quantity of objects. The results for the experiments are shown in table \ref{tab-transfer-comparison}.

Based on the results, when using the SFR with the pre-grasp in the real system, MDP-MOGT reduced the number of transfers by 6.38\% and the number of lifts by 9.26\% when compared to the naive approach. Similar results were observed when comparing the two transfer approaches using the grasping routine with the model and the finger flexion synergy. This showcases the superiority of the MDP-MOGT approach. 
Similarly, When comparing the MDP-MOGT results between the SFR with the pre-grasp and the grasping routine using the pre-grasps, finger flexion synergy, and the models, the routine with models outperformed the SFR routine by 6.81\% in the number of transfers and 14.28\% in the number of lifts in the real system. Similar results were observed in simulations as well. This showcases the improvement made because of the models and the finger flexion synergy.

\begin{table}[h]
\caption{Results for transfer approaches}
\label{tab-transfer-comparison}
\begin{center}
\begin{tabular}{|c|c|c|c|}
\hline
\textbf{\begin{tabular}[c]{@{}c@{}}Approach\end{tabular}} &
\textbf{\begin{tabular}[c]{@{}c@{}}Grasping\\routine\end{tabular}} &
\textbf{\begin{tabular}[c]{@{}c@{}} Average\\transfers\end{tabular}} &
\textbf{\begin{tabular}[c]{@{}c@{}} Average\\lifts\end{tabular}}\\
\hline
Naive & SFR (Real) & 4.7 & 5.4\\
\hline
MDP-MOGT & SFR (Real) & 4.4 & 4.9\\
\hline
Naive & Estimation models (Real) & 4.2 & 4.9\\
\hline
MDP-MOGT & Estimation models (Real) & 4.1 & 4.2\\
\hline
MDP-MOGT & SFR (Simulation) & 6.0 & 7.6\\
\hline
MDP-MOGT & Estimation models (Simulation) & 5.9 & 6.8\\
\hline
\end{tabular}
\end{center}
\end{table}

\section{Discussion \& Conclusion}

This paper proposed several new MOG techniques and two object transferring approaches that take advantage of MOG. The MOG techniques include clustered-probability-based pre-grasp, best expectation pre-grasp, maximum capability pre-grasp, and a data-driven deep learning model to predict the quantity of objects in a grasp after the hand lifts, when the hand is in the pile. The two object transferring approaches introduced is the naive transfer approach and the Markov-decision-process-based multi-object grasping transfer approach. 
We have evaluated the proposed strategies for transferring spheres in a simulation environment and the real system. The experiment results show that our approach outperforms single object transferring and the naive transfer approach. 
We also discovered that a large volume grasp does not translate to grasping a higher quantity of objects. 
In the future, we will test our approach on different shapes of objects and explore other deep neural network architectures to increase the precision of the models.

\section*{Acknowledgment}
This material is based upon work supported by the National Science Foundation under Grants Nos. 1812933 and 191004.

\bibliographystyle{IEEEtran}
\bibliography{main}

\end{document}